\documentclass[12pt]{article}

\usepackage[bitstream-charter]{mathdesign}
\usepackage[mathcal]{eucal}

\usepackage{float}
\usepackage{algorithm}
\usepackage{amsmath}
\usepackage{graphicx}
\usepackage{graphicx}
\usepackage{array}
\usepackage{tabularx}
\usepackage{color,epsfig}
\usepackage{cite}
\usepackage[caption = false,font = footnotesize]{subfig}
\usepackage{url}
\usepackage[colorlinks=true,citecolor=blue]{hyperref}
\usepackage{a4wide}
\usepackage[framemethod=tikz]{mdframed}
\usepackage[auth-sc,affil-sl]{authblk}

\definecolor{mycolor}{rgb}{0.122, 0.435, 0.698}

\newmdenv[innerlinewidth=0.5pt, roundcorner=4pt,linecolor=mycolor,innerleftmargin=6pt,
innerrightmargin=6pt,innertopmargin=6pt,innerbottommargin=6pt]{mybox}

\DeclareMathAlphabet{\baz}{OML}{cmm}{b}{i}

\def\b0{\mbox{\boldmath $0$}}

\floatstyle{ruled}
\newfloat{algorithm}{tbp}{loa}
\providecommand{\algorithmname}{Algorithm}
\floatname{algorithm}{\protect\algorithmname}

\begin{document}

\title{Conditional computation in neural networks: principles and research trends}

\author[1]{Simone Scardapane\thanks{Corresponding author. Remaining authors are listed in alphabetical ordering.}}
\author[2]{Alessandro Baiocchi}
\author[2]{Alessio Devoto}
\author[3]{Valerio Marsocci}
\author[4]{Pasquale Minervini}
\author[1]{Jary Pomponi}
\affil[1]{\small DIET Department, Sapienza University of Rome}
\affil[2]{\small DIAG Department, Sapienza University of Rome}
\affil[3]{\small Geomatics Research Group, KU Leuven}
\affil[4]{\small School of Informatics, University of Edinburgh}

\maketitle

\begin{mybox}
Please cite the published version as \textit{Scardapane, S. et al., 2024. Conditional computation in neural networks: principles and research trends. Intelligenza Artificiale, in press, pp. 1-16.} Feedback can be submitted to simone.scardapane@uniroma1.it.
\end{mybox}

\begin{abstract}
This article summarizes principles and ideas from the emerging area of applying \textit{conditional computation} methods to the design of neural networks. In particular, we focus on neural networks that can dynamically activate or de-activate parts of their computational graph conditionally on their input. Examples include the dynamic selection of, e.g., input tokens, layers (or sets of layers), and sub-modules inside each layer (e.g., channels in a convolutional filter). We first provide a general formalism to describe these techniques in an uniform way. Then, we introduce three notable implementations of these principles: mixture-of-experts (MoEs) networks, token selection mechanisms, and early-exit neural networks. The paper aims to provide a tutorial-like introduction to this growing field. To this end, we analyze the benefits of these modular designs in terms of efficiency, explainability, and transfer learning, with a focus on emerging applicative areas ranging from automated scientific discovery to semantic communication.
\end{abstract}

\section{Introduction}
\label{sec:introduction}
In the last twenty years, neural networks (NNs) have undergone two opposing trends. On one hand, the number of practical applications has continued to grow, fueled by successes in, among others, language modeling \cite{jiang2024mixtral}, drug design \cite{wang2023scientific}, rendering, and language-vision reasoning \cite{pfeiffer-etal-2022-xgqa}. On the other hand, their design has crystallized around a very small set of layers (e.g., multi-head attention) and principles (e.g., permutation equivariance), while the focus has shifted on scaling up their training, both in terms of data and parameters  \cite{jiang2024mixtral}. Apart from scale, maybe less than a dozen components and variations thereof are enough to categorize the vast majority of neural networks deployed nowadays.

Among these design principles, \textit{sequentiality} has remained a key component. NNs, be they convolutional, recurrent, or transformers, are composed of a stack of differentiable operations, which are activated in sequence for each input to be processed. Stated in another way, their \textit{computational graph}, i.e., the sequence of primitive operations executed on the underlying hardware, is fixed beforehand when instantiating them. Because NNs have continued to scale in depth and width, this has led to several issues in terms of computational performance and efficiency \cite{fournier2023practical}. Standard techniques to make NNs more efficient only address this partially, by replacing the original network with other \textit{static} models having {fewer} layers (distillation \cite{wang2021harmonized}, pruning \cite{Kim_2024_WACV}), less precision per parameter, or by approximating each weight matrix (e.g., via low-rank factorization).

By viewing neural networks as computing systems, this behavior is counter-intuitive. Hardware components are designed based on their expected peak usage, while only executing a fraction of their resources at any given time (e.g., memory). Similarly, software libraries and operating systems are composed of millions of lines of code, only a handful of which are selected and run in a given moment, thanks to the use of branches, loops, and conditional execution. Recently, a large body of literature has flourished on embedding similar \textit{sparse modularity} principles in the design of neural networks \cite{pfeiffer2023modular}. Based on the original idea of \textit{conditional computation} \cite{bengio2015conditional}, this has blossomed into a large number of practical implementations, ranging from mixture-of-experts (MoEs) \cite{shen2019mixture} to dynamic mechanisms to select tokens and layers in transformers \cite{meng2022adavit}. This tutorial is intended as an organized, uniform overview and entry point into this growing body of literature.

The benefits of developing networks that can dynamically adjust their computational graph {go} beyond memory or time efficiency. More and more, neural networks are treated in the same way as software \cite{kandpal2023gittheta}, and deploying them requires the possibility of quickly debugging their predictions \cite{bontempelli2022concept}, continually fine-tuning them on new data, and transferring parts of their knowledge from one network to the other \cite{ansell-etal-2022-composable}, similarly to standalone software libraries. As we will see, dynamically-activated neural networks provide a principled way to improve both zero-shot transfer and generalization \cite{ponti2022combining} (Sec. \ref{subsec:specialization}) and explainability of the models (Section \ref{subsec:explainability}). This aligns with the requirements of many novel applications of NNs, from scientific discovery \cite{wang2023scientific} to AI-native telecommunications \cite{xing2020earlyexit,strinati2024goal}. In particular, smart semantic communication networks \cite{strinati2024goal} envisioned in the so-called \textit{beyond 6G} model necessitate networks that can flexibly adapt to bandwidth and energy constraints, while ensuring transferability and communication through separate neural modules.

For the purpose of this tutorial, we consider three flavors of dynamism: (a) neural components that can restrict their computation to a smaller subset of input tokens ({\color{magenta}dynamic input sparsity}); (b) layers that can selectively activate sub-components for processing a token ({\color{teal}dynamic width sparsity}); and (c) layers that can be completely skipped during their execution ({\color{orange}dynamic depth sparsity}). As we show in Section \ref{sec:formulation}, a simple mathematical formalism encompasses all three cases. Section \ref{sec:formulation} also highlights how modularity can be achieved with the addition of a small set of primitives to our networks' toolkit, namely, the possibility of sampling in a differentiable way elements from a set. We discuss in Section \ref{subsec:gumbel_softmax} the simplest technique to this end, the Gumbel-Softmax trick \cite{jang2016categorical,maddison2016concrete}, and some common extensions.

We then proceed to discuss three notable implementations of these concepts in Section \ref{sec:implementation}: early-exit (EE) models, mixture-of-expert (MoE) layers, and token selection mechanisms. While these models are generally discussed in separate fashions (e.g., see \cite{scardapane2020should,matsubara2022split} for EEs, and \cite{yuksel2012twenty,fedus2022review} for MoEs), viewing them as specific instances of a general framework highlights many similar trends and characteristics. In fact, we argue that designing networks with dynamically activated components is not simply a matter of enhancing performance: in all these cases, the resulting networks are more apt at adapting to variable system's constraints (e.g., decreased energy usage),  specialization, catastrophic forgetting, and multimodality. We build on these insights in Sections \ref{sec:research_directions} and \ref{sec:conclusions}, where we list potential research directions for these models including adapting their computational cost and energy at inference and training time in an elastic way, zero-shot transfer, robustness, and explainability.

\textbf{Relation to prior works}: we do not claim to be the first to discuss these ideas, nor to present them in a general way. We simply hope to provide a simple, cohesive entry point into an extremely fascinating and growing research direction in the literature. Recently, modular deep learning \cite{pfeiffer2023modular} has been proposed as a general term for neural networks where computation is functionally decomposed into units that are sparsely activated. Compared to \cite{pfeiffer2023modular}, our formalism is simpler and we focus on a smaller set of ideas: for example, we do not consider input composition methods, nor soft routing strategies, and we focus on viewing modularity as discrete sampling inside neural networks. Hence, we provide an orthogonal view to \cite{pfeiffer2023modular}, to which we refer for a broader outlook on modularity and composition. We also focus on a tutorial exposition, preferring simplicity and clarity to completeness. We only describe the simplest techniques for differentiable sampling, and we refer to \cite{niculae2023discrete} for a larger and more in-depth introduction to this field. Additional entry points in the literature include \cite{han2021dynamic} (for an overview up to 2021 of dynamism in NNs) and \cite{fournier2023practical} for designing faster and more efficient transformer models.

\section{Three types of conditional computation}
\label{sec:formulation}
\subsection{A general formalism for sparse modularity}
Neural networks can be described as the composition of several trainable, differentiable operations of the form:
\begin{equation}
    y = f(x, w)
    \label{eq:layer}
\end{equation}
where $x$ denotes the input, $w$ the module's parameters, and $f$ the specific operation. Because of functional composition, $f$ can represent an operation at any level of complexity, e.g., a basic linear algebra operation, a layer (convolutional, recurrent, ...), or a composite block, such as the classical combination of token mixing, channel mixing, and layer normalization operations found in transformers. 

Most of our discussion focuses on transformer-like architectures, where the input is composed of a set of \textit{tokens}, representing parts of the complete input \cite{riquelme2021scaling} (e.g., subwords in a sentence or patches in an image). The input $x$ in \eqref{eq:layer} is purposefully left vague: depending on the scenario and the layer, $x$ can be a single token (e.g., for the fully-connected blocks inside a transformer), the entire set of tokens, a mini-batch of inputs, or other forms of data aggregation.

We are interested in augmenting \eqref{eq:layer} to allow the dependence of the different quantities to vary based on some \textbf{conditioning input} $x_c$, where $x_c$ can be equal to $x$, some linear or non-linear projection of $x$ (in order to decouple the conditioning input from the input used for the layer and increase the flexibility of the dynamic mechanism), or some additional input information (e.g., a latent token representing the language of the speaker in an audio model \cite{lin-etal-2021-learning}).

\begin{figure}
\centering
\includegraphics[width=0.7\textwidth]{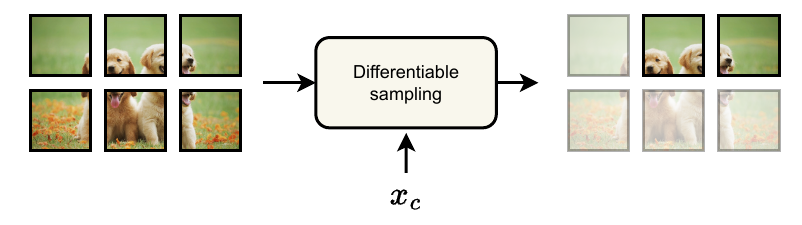}
\caption{{\color{magenta}Input sparsity:} A differentiable mechanism subsamples input tokens to be processed by the later parts of the network (we show original image patches in the figure, but the tokens can be equivalently be replaced by their latent representations if we consider an intermediate layer of the architecture).}
\label{fig:input_sparsity}
\end{figure}

Suppose we have available some trainable \textit{subsampling} operation, defined over the conditioning input $x_c$ and over some set $\mathcal{S}$, such that:
\begin{equation}
    \Gamma(\mathcal{S}, x_c) \subseteq \mathcal{S} \,.
    \label{eq:subsampling}
\end{equation}
Practically, sets are always represented in some matrix format inside neural networks (e.g., by stacking all elements row-wise), in which case we assume the relevant quantities in \eqref{eq:subsampling} to follow the same conventions. As an example, if $\mathcal{S}$ is represented by a matrix $\mathbf{S}$, \eqref{eq:subsampling} can be implemented by row-wise masking of the corresponding elements:
\begin{equation}
    \Gamma(\mathbf{S}, x_c) = \mathbf{M} \odot \mathbf{S}
    \label{eq:matrix_masking}
\end{equation}
where $\mathbf{M}$ is a binary mask of suitable shape with rows set to either $\mathbf{0}$ or $\mathbf{1}$. Different implementations can be obtained depending on whether the size of the output mask is known in advance (e.g., how many tokens to keep in a given layer), or must be estimated by $\Gamma$ itself. We will see later on in Section \ref{subsec:gumbel_softmax} how such an operation can be implemented in practice.

This small addition allows us to implement several levels of modularity inside our neural network. First, assume the input $x$ can be decomposed into smaller components $x = \left\{x_1, \ldots, x_n\right\}$. In many cases, this is a natural decomposition, e.g., tokens inside a transformer, time instants for a sequence, or frames for a video input. However, they can also correspond to additional register tokens \cite{darcet2023vision}, to elements extracted from some external memory (e.g., AdaTape \cite{xue2023adaptive}), or additional views or modalities. Then, {\color{magenta}dynamic input sparsity} (Fig. \ref{fig:input_sparsity}) is achieved by combining the layer with a subsampling operation on the input:
\begin{equation}
    \text{Input sparsity}: f(\Gamma(x, x_c), w)
\end{equation}
This allows the layer to focus exclusively on input components that are relevant to the current operation. As an example, consider an image with a very wide, uniformly blue background: for the majority of tasks, we can imagine the layer to be able to operate even when removing the vast majority of tokens corresponding to such background \cite{wojcik2023adaptive}. This is, indeed, observed in practice, as we'll describe in Section \ref{subsec:token_sampling}.

\begin{figure}
\centering
\includegraphics[width=0.6\textwidth]{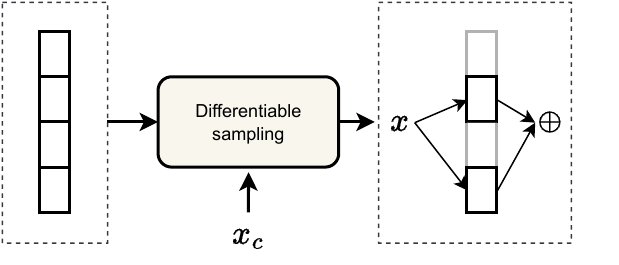}
\caption{{\color{teal}Width sparsity:} Different parts of a layer (e.g., \textit{experts}) can be activated based on the conditioning value.}
\label{fig:width_sparsity}
\end{figure}

Second, assume the weights themselves can be decomposed into blocks $b = \left\{w_1, \ldots, w_m\right\}$, and the entire function decomposed as:
\begin{equation}
    f(x, w) = \oplus(\left\{f_1(x, w_1), \ldots, f_m(x, w_m)\right\}),
    \label{eq:function_decomposition}
\end{equation}
\noindent where $\oplus$ is a permutation equivariant operation that aggregates the results of the different blocks. This is also a very natural decomposition in practice: for example, $f_i$ could correspond to a single output channel in a convolutive layer, with $\oplus$ being a concatenation. Or, $f_i$ can be a head in a multi-head attention layer, with $\oplus$ being a concatenation followed by a projection. In this case, what we call {\color{teal}dynamic width sparsity} (Fig. \ref{fig:width_sparsity}) can be achieved by subsampling the blocks:
\begin{align}
    \text{Width sparsity}: f(x, \Gamma(w, x_c)) = \nonumber \\ \oplus\left( \Gamma(\left\{ f_1(x, w_1), \ldots, f_m(x, w_m), x_c \right\} \right)
    \label{eq:width_sparisty}
\end{align}
Imposing some form of structure over $w$ is fundamental since subsampling each weight separately would result in a highly unstructured form of dynamic pruning, which is generally very difficult to optimize and control \cite{jaszczur2021sparse}. Once again, this formulation is very generic: for example, if each $f_i$ corresponds to an activation function, subsampling with cardinality $1$ corresponds to choosing the best activation function for each token or each input, in a form of dynamic model selection mechanism \cite{liu2018darts}. If each $f_i$ is an entire model, this can be used to route information across blocks with different types or complexity (see, e.g., layer stitching \cite{pan2023stitchable}). We will see in Section \ref{subsec:moes} a common implementation of this principle in the context of MoEs, and defer more general discussions to Section \ref{sec:research_directions}.

\begin{figure}
\centering
\includegraphics[width=0.7\textwidth]{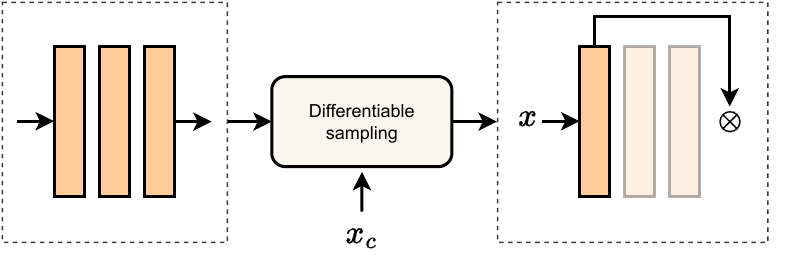}
\caption{{\color{orange}Depth sparsity:} A subset of the layers can be deactivated by the sampling mechanism, like in early-exit networks.}
\label{fig:depth_sparsity}
\end{figure}

Third, even if a weight decomposition is unavailable, dynamic computation can still be achieved by considering the entire layer as a single block and conditionally skipping its execution. This {\color{orange}dynamic depth sparsity} (Fig. \ref{fig:depth_sparsity}) can be achieved easily by rewriting the layer with an additional, untrainable scalar weight $\sigma = 1$, and writing:
\begin{align}
   \text{Depth sparsity}: f^\prime(x, w, \Gamma(\left\{\sigma\right\}, x_c)) = \\ \sigma(x_c) f(x, w)  + (1-\sigma(x_c))x
   \label{eq:depth_sparsity}
\end{align}
where we assume the empty set $\emptyset$ to be implemented as $0$ in practice. As a variant of this idea, we can skip entire blocks of layers by \textit{exiting} the network instead of adding a skip connection in \eqref{eq:depth_sparsity}. We will see concrete implementations of this concept in Section \ref{subsec:early_exits}.

\subsection{Discrete sampling: the Gumbel-Softmax trick}
\label{subsec:gumbel_softmax}
Before proceeding, we discuss briefly the implementation of the subsampling layer $\Gamma$. As a prototypical example, we restrict our analysis to subsampling a single element from the set in input. In this case, a very common implementation is the so-called Gumbel-Softmax (GS) trick \cite{jang2016categorical}, also known as the concrete distribution \cite{maddison2016concrete}. In this section, we only provide a high-level overview, and we refer to \cite{huijben2022review} for a fuller exposition.

\begin{figure*}
\centering
\includegraphics[width=0.9\textwidth]{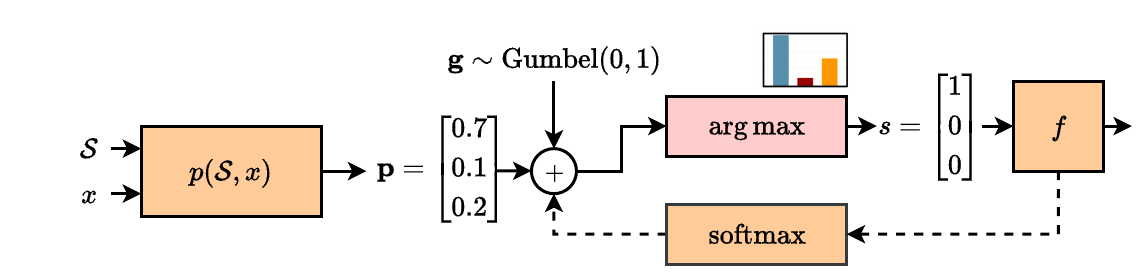}
\caption{Overview of the Gumbel-Softmax trick. We show in orange differentiable operations (the argmax's gradient being zero almost everywhere) and with {a dashed} arrow the relaxed backward path.}
\label{fig:gumbel_softmax}
\end{figure*}

First of all, we process the inputs with some trainable layer $p(\mathcal{S}, x) \in \mathbb{R}^{\lvert \mathcal{S} \rvert}$ to provide a real-valued score for each element in $\mathcal{S}$, which is proportional to its probability of being sampled. In this section, we drop the subscript from the conditioning input $x_c$ for readability. The implementation of $p$ depends on the use case. For example, suppose that the conditioning input is a tensor $x \in \mathbb{R}^{h \times w \times d}$ coming from intermediate output of a CNN having width $w$, height $h$, and $d$ channels, while $\mathcal{S}$ is the set of output channels in a convolutive layer, of which we want to select a single one \cite{herrmann2020channel}. We associate to each channel a trainable vector $\mathbf{c}_i$, and compute an affinity score via the dot product with a pooled representation of the image \cite{herrmann2020channel}:
\begin{align}
    \mathbf{v} = \text{MLP}\left(\sum_{i,j} x_{ij} \right) \\
    p(\mathcal{S}, x) = \left[\mathbf{v}^\top\mathbf{c}_1, \ldots, \mathbf{v}^\top \mathbf{c}_{\lvert \mathcal{S}\rvert} \right]
\end{align}
Many other designs for $p$ are possible based on $x$ and $\mathcal{S}$. Let us denote by $\mathbf{p}$ the output of $p$, and by $g_i$ samples from the so-called Gumbel distribution \cite{jang2016categorical}. We also assume that the elements of $\mathcal{S}$ correspond to integers, $\mathcal{S} = \left\{1, \ldots, \lvert \mathcal{S}\rvert\right\}$, in which case we simply write $\Gamma_{\lvert \mathcal{S}\rvert}(x)$ for the subsampling over the set of integers. Then, the following operation provides an unbiased sample from $\mathcal{S}$, with the probability of sampling the $i$-th element $p(s \in \mathcal{S}) \propto \exp(p_s)$ proportional to $p_s$:
\begin{equation}
    s = \underset{i}{\arg\max} \;\;\left\{ p_i + g_i \right\} \,.
    \label{eq:gs_hard}
\end{equation}
Removing the Gumbel noise $g_i$ {corresponds} to taking the element with {the highest} score $p_i$, while sampling provides a degree of freedom that is helpful in exploring possible alternatives. Since the probabilities $p_i$ are implicitly trained via $p(\mathcal{S}, x)$, the network can learn to select the element from $\mathcal{S}$ which is most useful for the specific input $x$. In the case of channel selection, for example, this can lead to specializing single channels to specific types of inputs.

The operation in \eqref{eq:gs_hard} is not easily trainable via gradient descent (since its gradient will be zero almost everywhere), but it can be relaxed with a softmax approximation:
\begin{equation}
    s_i = \frac{\exp(\left(p_i + g_i\right)/\tau)}{\sum_j \exp(\left(p_j + g_j\right)/\tau)}  \,.
    \label{eq:gs_soft}
\end{equation}
The quality of the approximation can be controlled by the user-defined parameter $\tau$, sometimes called the temperature \cite{huijben2022review}. The values in $\mathbf{s}$ are not binary anymore, being generic convex combinations of the (one-hot) representations from the elements in $\mathcal{S}$. However, a common relaxation is to use the binary values in \eqref{eq:gs_hard} during the forward pass of the network, and relax the gradients to use the soft approximation in \eqref{eq:gs_soft} during the backward pass. This is called straight-through estimation (STE). A high-level overview of the complete schema is given in Fig. \ref{fig:gumbel_softmax}.

The GS trick can be easily extended to sampling more than a single value by replacing the $\arg\max$ with a top-k operation \cite{kool2020ancestral}. Suitable generalizations, such as the entmax family \cite{correia2019adaptively}, can also sample binary vectors with a variable number of elements. The simplicity of the GS trick makes it widespread in many applications, but several other types of sampling layers can be found in the literature, especially for more complex combinatorial spaces, for which we refer to \cite{niculae2023discrete}. A larger overview of possible sampling operations for MoE layers is also given in Sec. \ref{subsec:routing}.

\section{Concrete implementations}
\label{sec:implementation}

The previous section has introduced a generic framework for designing networks with dynamic computational graphs. In the literature, these ideas have coalesced around a few, notable set of implementations. In this section we overview three of the most popular ones: \textbf{early exits} (Sec. \ref{subsec:early_exits}), \textbf{MoE layers} (Sec. \ref{subsec:moes}), and \textbf{token sampling} (Sec. \ref{subsec:token_sampling}) as examples of networks having dynamic {depth, width and input}, respectively. In all cases we focus on {the key} strengths and drawbacks of each approach, highlighting how they connect to the framework of Sec. \ref{sec:formulation}.

\subsection{Early exits}
\label{subsec:early_exits}

Early exit neural networks (EENNs) were introduced based on the idea that not all the inputs to a network {are required} to go through all the layers of the model to be correctly classified \cite{teerapittayanon2016branchynet}. In fact, the accuracy of a network can decrease with respect to the depth on particularly easy samples, a phenomenon known as \textit{overthinking} \cite{sarti2023}.

Consider a neural network (sometimes called the backbone network in this context), which can be either trained from scratch or pre-trained. In an early-exit framework, the backbone is augmented with auxiliary classifiers (early exits) which are connected at intermediate outputs of the backbone. These auxiliary blocks can be trained all at the same time or in a layer-wise fashion \cite{scardapane2020should}, while at inference time they are used to halt the computation when the model is confident enough about the prediction. This is an example of dynamic depth sparsity (Section \ref{sec:formulation}), except that an entire subset of layers is skipped at the same time.

We introduce first the most common formulation of EENNs, in which early exits are jointly trained while the early exit logic is defined manually with additional hyper-parameters. As already stated, we can see a neural network as a composition of $b$  sequential blocks:
\begin{equation}
    f(x) = \mathrm{f}_{b} \circ \mathrm{f}_{b-1} \circ \cdots \circ \mathrm{f}_{1}(x) 
\end{equation}
\noindent where $\circ$ denotes function composition, $b$ is the last layer of the backbone, and we remove the dependency on the parameters for simplicity. Where to place the auxiliary classifiers is in general {a hyper-parameter} \cite{scardapane2020should}, but here, for clarity, we add a classifier after each block, producing multiple prediction functions: 
\begin{equation}
    y_i(x) = \mathrm{c}_{i} \circ \left( \mathrm{f}_{i} \circ \mathrm{f}_{i-1} \circ \cdots \circ  \mathrm{f}_{1}(x) \right)
\end{equation}
where $c_i$ is a small classifier (e.g., for a CNN, a typical design is to have a global average pooling operation followed by a fully-connected layer, while for a transformer we can apply a fully-connected layer on a specialized class token). This formulation leads to an EENN which is capable of classifying input samples at any stage of the computation through the sequence $y_1(x), y_2(x), \ldots, y_{b-1}(x), f(x)$. Ideally, each classifier can specialize on a subset of the training samples based on the complexity of the input. In the simplest case, given a desired output $y$, all early exits can be trained simultaneously by minimizing the sum of the losses for each classifier:
\begin{equation}
    L = \alpha \text{CE}(y, f(x)) + \sum_{i=1}^{b-1} \beta_i \text{CE}(y, y_i(x)),
\end{equation}
\noindent where CE is the cross-entropy loss, and $\alpha$, $\beta_1$, $\ldots$, $\beta_{b-1}$ are hyper-parameters that balance the loss contribution from each early exit. At inference time, an exit can be chosen by comparing, e.g., the predicted class probability or the entropy of the prediction to a user-defined parameter acting as a threshold. Other strategies include combining predictions based on trainable \cite{scardapane2020differentiable} or geometric \cite{wolczyk2021zero} ensembling.

These networks have found applications in many fields recently. Concerning NLP, many strategies with BERT have been proposed \cite{zhou2020bert, xin2020deebert, zhu2021leebert}, varying from patience-based EE \cite{zhou2020bert} to extra classification layers \cite{xin2020deebert}. For computer vision, EE applications vary based on the strategy, including multi-exiting \cite{Bakhtiarnia2021improving, kouris2022multi, wang2021harmonized, bolukbasi2017adaptive}, and loss-weight adjustment \cite{wang2019dynexit}. Also, the tasks vary, spanning semantic segmentation \cite{kouris2022multi}, video recognition \cite{ghodrati2021frameexit}, adversarial training \cite{karpikova2023fiancee}, and image compression \cite{xing2020earlyexit}.
Recently, EE strategies have been proposed also for vision-language models \cite{tang2023you}.

Although this formulation of EENNs (or minor variations thereof) is very common in the literature (see e.g., \cite{teerapittayanon2016branchynet,bolukbasi2017adaptive,zhou2020bert,matsubara2022split,schuster2022confident,tang2023you}), it only fits partially our view of dynamic sparsity, since the computational graph is unchanged during training and the dynamism is only obtained via an external heuristic. To this end, \cite{scardapane2020differentiable} proposed a mechanism to optimize the early exit selection during training, called \textit{differentiable branching}. Suppose we augment each early exit with an additional output that we denote as $\gamma_i(x)$. The key idea of differentiable branching is to define a new recursive output, which is given by a soft composition of the original outputs:
\begin{equation}
    \tilde{y}_i(x) = \gamma_i(x) y_i(x) + (1-\gamma_i(x))\tilde{y}_{i+1}(x)
\end{equation}
where the recursion stops at $i=b$, where $\gamma_i(x)=1$ by definition. At inference time, this can be turned into an early exit module by viewing $\gamma_i(x)$ as a binary classifier and thresholding it at $0.5$. By replacing the gates $\gamma_i(x)$ with samples from a GS distribution, we return to the general setting described in Section \ref{sec:formulation}. In fact, if the outputs of $\gamma_i(x)$ are binary we can identify:
\begin{equation}
    \Gamma_{b-1}(x) = \begin{bmatrix} \gamma_1(x) \\ (1-\gamma_1(x))\gamma_2(x) \\ \vdots \\ \left[\prod_{i=1}^{b-2} (1-\gamma_i(x))\right]\gamma_{b-1}(x) \end{bmatrix}
    \label{eq:Gamma_early_exit}
\end{equation}
with the subsampling block over the indexes of the early exits. This can also be given a Bayesian interpretation under the stick-breaking process \cite{pomponi2021probabilistic}. While the sampler in \eqref{eq:Gamma_early_exit} is conditioned on the full input, other choices are possible. For example, in multitask learning a separate task embedding can be used to select different exits for different tasks \cite{zhang2022unified}, while in the case of transformers, only the class token can be used to decide an optimal exit strategy \cite{He2021}. Being able to train the choice of early exit is fundamental because it allows us to regularize it further for, e.g., better inference-accuracy trade-offs: we revisit this idea in Section \ref{subsec:efficiency}. See \cite{scardapane2020differentiable,matsubara2022split} for in-depth reviews on EENNs and \cite{scardapane2020differentiable} for a fuller description of differentiable branching.

\subsection{Mixture-of-experts}
\label{subsec:moes}


Mixture-of-Experts (MoEs) models were introduced more than thirty years ago as adaptive algorithms to perform dynamic ensembles of individual machine learning algorithms, denoted as \textit{experts} in that context \cite{yuksel2012twenty}. Recently, MoE layers in neural networks have gained popularity, especially for LLMs and large vision modules \cite{riquelme2021scaling}. They allow scaling the model's parameters while keeping the compute budget constant, and they offer the possibility of distributing the training by decomposing the computation of a layer over multiple GPUs \cite{riquelme2021scaling}. As a side product, they offer dynamic width sparsity by selectively activating only parts of the layer for each input. At the time of writing, the largest open source language and vision models are based on a MoE formulation \cite{lepikhin2020gshard,fedus2021switch,riquelme2021scaling,dai2024deepseekmoe,jiang2024mixtral}.

In a MoE layer, the trainable parameters are grouped into topologically identical blocks called \textit{experts}. In the forward pass, a decision block (called \textit{router} in this context) selects a subset of expert blocks (typically, only a fixed number of $k$ experts, with $k$ selected beforehand, is chosen) that will be activated and routes the input data to the selected experts. More formally, the MoE layer can expressed as:
\begin{equation}
    f(x, w) = \sum_{i=1}^n \gamma_i(x) f_i(x, w_i) 
    \label{eq:moe}
\end{equation}
where $\gamma$ is the routing function and $f_1, f_2 \dots f_n$ are the experts. In its simplest form, this is a sparse linear combination of the outputs of the experts. This is immediately seen to be an instance of \ref{eq:width_sparisty} with $\Gamma_n(x)=\left[\gamma_1(x), \gamma_2(x), \ldots, \gamma_n(x)\right]$, provided that the outputs of $\gamma_i(x)$ are sparse.

The unique advantage of MoEs is that only a fraction of the network parameters, depending on the chosen value for $k$, are used for computation. Such fraction can be tuned by changing the value of $k$. Additionally, one can devise routing functions such that experts focus on different areas of the input space, leading to specialized sub-networks (see Sec. \ref{subsec:specialization} for a discussion on specialization and Sec. \ref{subsec:routing} for an overview of routing algorithms for MoE layers).

As stated earlier, the seminal idea of MoEs dates back to \cite{jordanMoE}, where this paradigm is regarded as an ensembling technique with soft gates. The first work that adopts MoEs in a modern deep learning scenario is \cite{shazeer2017outrageously}. Here, the gate becomes sparse, i.e., a fixed top-$k$ operation and the main goal is scaling the number of parameters. 
Based on these initial results, in the past few years, the deep learning community witnessed a steep growth of MoE's success in {conjunction} with transformer architectures. 

Unlike early exiting, where the routing function typically works at the level of input samples (e.g., images for a CNN), {an MoE} layer as in \eqref{eq:moe} can work at the level of individual tokens in a transformer. More specifically, in the fully-connected layers of a transformer block, each token can be assigned to one or more experts. Transformer MoEs have been proposed for NLP \cite{fedus2021switch, puigcerver2020scalable}, for vision \cite{riquelme2021scaling} and even for multimodal training \cite{mustafa2022multimodal}. Additional variants can be applied to individual heads of an attention block \cite{zhang2022mixture} or adapters in a fine-tuning context \cite{zadouri2023pushing}.

\subsection{Routing in Mixture of Experts} 
\label{subsec:routing}

Different routing strategies can be adopted to orchestrate the experts. These are defined in the gating function $\gamma$ and usually leverage some differentiable sampling method similar to the Gumbel-Softmax trick (Sec. \ref{subsec:gumbel_softmax}). Routing is a crucial aspect of any MoE architecture, as it must perform a matching between input data and expert network, conditioned on the input data itself. Decisions taken by the routing function govern the training. They can determine faster convergence and expert specialization or cause \textit{expert starvation}, a peculiar situation where only a few experts are always activated due to their ever-increasing expertise.

The routing algorithm usually adopts a dot product for computing similarity between the embedding of each routed sample (e.g. a token in a transformer) and $n$ trainable expert embeddings (one for each expert). Once this operation has occurred, the assignment can happen according to various methods. 

In \emph{top-$k$ token choice}, each token is assigned to the $k$ most similar experts, where $k$ is an arbitrary value. This is the original approach from \cite{shazeer2017outrageously}, and it usually requires additional regularization to avoid expert starvation and training instability. The same approach has been adopted with minor changes (e.g. balancing loss and the $k$ factor) in \cite{rajbhandari2022deepspeedmoe,fedus2021switch}. \emph{Top-$k$ expert choice}, proposed in \cite{zhou2022mixture}, tries to solve the expert balancing problem by assigning each expert to the $k$ most similar tokens, where $k$ is an arbitrary value. Other works \cite{clark2022unified, rosenbaum2017routing} cast the expert assignment problem into a \emph{reinforcement learning} framework and train over a routing policy to assign each token to a single expert. Finally, \cite{zuo2022taming} achieves surprisingly good results by randomly assigning experts to routed samples. An empirical comparison of different routing functions can be found in \cite{liu2024routers}, while a broader overview of MoE layers is given in \cite{fedus2022review}.

\subsection{Soft routing}
\label{subsec:soft_moe}
We discuss here a variant of the MoE layer in \eqref{eq:moe}, known as \textit{soft} MoE, which has become popular recently to avoid training instabilities in the layer. The key idea is to use the gating function to combine the layer's inputs \cite{puigcerver2023sparse} or weights \cite{muqeeth2023soft} instead of the model's outputs. These variants also provide a fixed compute budget which is decoupled from the total number of parameters, but they are generally easier to implement {and train}.

First, denote by $x_1, \ldots, x_n$ the tokens in input to the layer. The first soft MoE variant we consider can be defined as \cite{puigcerver2023sparse}:
\begin{equation}
    f_i(x, w_i) = f_i\left({\color{cyan} \sum_{i=1}^n \gamma_i(x_i) x_i}, w_i\right)
    \label{eq:soft_moe_1}
\end{equation}
i.e., each expert is activated with a soft combination of input tokens. Compared with standard top-$k$ routing, each expert is only activated once in this formulation, irrespective of the number of tokens. The output of the layer for a single token is then given by another soft combination, this time over the experts' outputs:
\begin{equation}
    f(x_i) = \sum_{i=1}^n \gamma_i(x_i) f_i(x, w_i)
\end{equation}
By comparison, another soft MoE variant can be achieved by performing a soft combination of the \textit{experts' weights} \cite{muqeeth2023soft}. Denote by $f(x,w)$ the generic architecture of a single expert, we have:
\begin{equation}
    f(x) = f\left(x, {\color{cyan} \sum_{i=1}^n \gamma_i(x) w_i} \right)
\end{equation}
A thorough comparison of the relative benefits of hard vs. soft routing is still lacking in the literature, but we highlight a few benefits in Sec. \ref{sec:research_directions}.

\subsection{Token dropping and token merging}
\label{subsec:token_sampling}

We conclude with a brief analysis of token selection mechanisms. In particular, we consider two cases of token selection, which we refer to as \textit{token dropping} (removal of tokens from the model) and \textit{token merging} (combination of tokens). Both are conditional computation techniques that can be employed in any transformer model. The fundamental idea behind token dropping and token merging is that the input data often contains information that is nearly useless for the final task. Since transformers operate on set-based inputs with no fixed cardinality, one can dynamically reduce the number of tokens being passed to each layer depending on the relevance of said tokens for the final task.

More formally, let $\mathbf{X}$ be the set of $n$ tokens inside a transformer model stacked horizontally. For the purpose of {this section,} the subsampling operation $\Gamma$ can be implemented as a matrix multiplication of the input matrix $\mathbf{X}$ and a mask $\mathbf{M}$:
\begin{equation}
\mathbf{X}^\prime = \Gamma_n(\mathbf{X}) = \mathbf{M}\mathbf{X}
\label{eq:subsampling_2}
\end{equation}
Depending on the properties of the matrix $\mathbf{M}$, we can obtain either token dropping or token merging:
\begin{itemize}
    \item Token dropping: $\mathbf{M}$ is a matrix of shape $n^\prime \times n$ that selects a subset of $n^\prime$ elements from the original tokens. This can be achieved by setting each row to a one-hot vector.
    \item Token merging: $\mathbf{M}$ is a binary matrix of shape $n^\prime \times n$, with the requirement that each column sum to one. In this way, the output tokens are combinations of the input ones, so each input token participates in a single output token. This can be seen as a dynamic variant of the standard pooling used in CNNs.
\end{itemize}
Both methods aim to reduce the computational cost of the forward pass of a transformer by reducing the number of tokens. The mask $\mathbf{M}$ is usually computed depending on the input data, $\mathbf{M} = \text{MLP}(\mathbf{X})$, in such a way as to eliminate only redundant information, such as background patches in an image. This is usually done either by small modules that are added to the backbone model \cite{rao2021dynamicvit} or by evaluating some channels added to the tokens for this purpose \cite{meng2022adavit}. For token selection, when dealing with mini-batches of inputs, the sub-sampling operation in \eqref{eq:subsampling_2} cannot be performed easily in a vectorized way. In these cases, \eqref{eq:subsampling_2} can be replaced by a masking operation as in \eqref{eq:subsampling}, with the additional constraint that subsequent selection modules can only select values that were not masked previously \cite{rao2021dynamicvit}. In these cases, performance gain only materialize at inference time, while at training time most operations are still executed in a masked way.

Token dropping and token merging have been largely explored in recent years; some of the most impactful works on these topics in the field of vision transformers are \cite{meng2022adavit,rao2021dynamicvit} for token dropping, while \cite{bolya2022token, pan2022less} explored token merging. These techniques have also been employed in other fields where transformers are predominant: in 3D computer vision token dropping has been applied to point cloud transformers by \cite{yang2019modeling}, while in NLP \cite{hou2022token} applied token dropping in the pre-training of BERT, obtaining significant speedups. 

\section{Research directions}
\label{sec:research_directions}

Designing neural networks with sparse modularity principles has a number of benefits. The first and most studied is increased efficiency, both in training and in inference (Sec. \ref{subsec:efficiency}). However, modular networks show emergent properties also in terms of specialization and transferability (Sec. \ref{subsec:specialization}), as long as providing a blueprint for a new type of explainability techniques (Sec. \ref{subsec:explainability}). We briefly overview these aspects in the following sections.

\subsection{Efficiency in training and inference}
\label{subsec:efficiency}

\begin{figure}
\centering
\includegraphics[width=.5\linewidth]{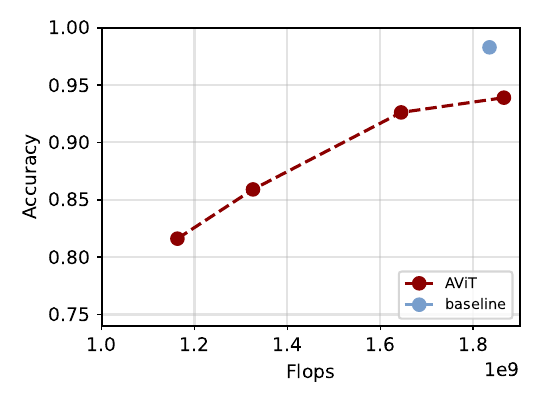}
\caption{{Example of accuracy-flops trade-off for inference with A-ViT \cite{yin2022vit}. Specifically, the architecture is a DeiT, trained on Imagenette.}}
\label{fig:token_halting}
\end{figure}

Conditional computation has gained significant attention for accelerating training and inference in deep learning models \cite{fournier2023practical, pfeiffer2023modular}. {Often, the time saving comes with a drop in accuracy, making it important to assess the accuracy/computation trade-off. An interesting example is shown in Fig. \ref{fig:token_halting}}. The approaches can vary based on several factors. For example, some research investigated the sparsification of CNNs \cite{Verelst_2020} and of transformers \cite{jaszczur2021sparse, NEURIPS2021_a61f27ab}, some others focused on some specific aspects of the net, such as the gradients \cite{dettmers2019sparse}, the backpropagation, the activations \cite{Chen_2023_CVPR} and the attention layers \cite{treviso2022predicting, child2019generating}. For clarity, we focus here{, mainly,} on efficiency aspects related to the three families of models seen in Sec. \ref{sec:implementation}, focusing on emerging trends and open challenges. {To make the analysis more timely and precise, we have reported all the methods analyzed in Table \ref{tab:eff}.}

When training from scratch an EENN, the joint training described in Sec. \ref{subsec:early_exits} cannot provide faster training since all exits must be trained simultaneously. Soft combinations of the early exits \cite{scardapane2020differentiable} combined with sampling tricks could be useful, but this aspect has been scarcely explored in the literature, possibly due to training instabilities and collapse. Still, EEs can provide indirect benefits to training by accelerating the convergence of deep neural networks \cite{bolukbasi2017adaptive, teerapittayanon2016branchynet}. They can also improve the inference efficiency of large-scale pre-trained models \cite{han2022learning, zhou2020bert}, making them more discriminative \cite{teerapittayanon2016branchynet}, act as a regularization technique \cite{sarti2023}, and possibly reduce training problems (e.g. vanishing gradient phenomena) \cite{passalis2020}. In addition, they can be trained in a layerwise fashion if one starts from a pre-trained network \cite{han2021dynamic}. {In Table \ref{tab:eff}, we can see that most of these methods require additional parameters, leading to a latency reduction.}

Compared to EEs, MoEs were investigated mostly to {speed up} training by distributing the experts across different GPUs \cite{fedus2022review}, or to accelerate inference by activating a subset of experts for each forward pass. For the former, tiling the model across separate machines requires customized implementations \cite{rajbhandari2022deepspeedmoe}. In addition, to achieve an effective stable throughput, training must avoid collapses of the routing function, and care must be taken to balance the number of tokens {that are} sent to the different experts. This last point can be achieved with the addition of so-called \textit{balancing losses}, e.g., batch prioritized routing \cite{riquelme2021scaling,clark2022unified, fedus2022review}. Due to these problems, open-source implementations of large MoE models have generally lagged behind proprietary models, with some preliminary steps being taken in this direction \cite{xue2024openmoe}{, as observable also in Table \ref{tab:eff}}. 

The benefits of applying MoEs can also vary depending on the specific component {that is} being replaced. In \cite{shazeer2017outrageously}, MoEs replace MLP layers to scale-up transformers. A similar methodology, for vision, was presented in \cite{riquelme2021scaling}. Other approaches proposed to replace attention layers \cite{zhang2022mixture}, entire blocks \cite{tan2023sparse} of the net, and adapters \cite{zadouri2023pushing}. By comparison, soft MoEs (Sec. \ref{subsec:soft_moe}) can reach similar gains as standard MoE layers while being simpler to train. Finally, the majority of MoEs are trained from scratch, but recently \textit{moefication} has emerged as an interesting research direction, in which a pre-trained model is converted to {an MoE} variant by clustering the activations \cite{qiu2023emergent,zhang2021moefication} {(see Table \ref{tab:eff} to see the approaches based on pre-training)}. These variants provide different accuracy-time trade-offs based on the specific routing function. Additional emerging research trends are exploring \textit{dynamic} variants of routing to provide a flexible inference budget instead \cite{piorczynski2023exploiting,wojcik2023adaptive}.

Finally, reducing the number of tokens with token selection techniques is a straightforward strategy to improve the inference time of the network. Most methods in this sense have focused on token dropping \cite{haurum2023tokens, yin2022vit, bolya2022token}: {A-ViT adopts a token halting approach (shown also in Fig. \ref{fig:token_halting});}AdaViT \cite{meng2022adavit} proposes a light-weight decision network, attached to the backbone to produce decisions on-the-fly;
DynamicViT \cite{rao2021dynamicvit} proposes an attention-masking strategy to block interaction of redundant tokens; MSViT \cite{Havtorn_2023_ICCV} proposes a conditional gating mechanism that selects the token scale for every image region; GTP-ViT \cite{Xu_2024_WACV} introduces a Graph-based Token Propagation (GTP) {to propagate the information of less significant tokens}. During training, token selection is generally achieved by masking the corresponding input elements (Sec. \ref{subsec:token_sampling}). Designing token selection mechanisms that can improve training time (for a fixed compute budget) is still an open challenge.

Some works have focused more on specific tasks, such as diffusion \cite{bolya2023token}, long-input sequences \cite{ainslie2023colt5}, segmentation \cite{liu2024revisiting}, or to better understand the general pattern of token dropping \cite{haurum2023tokens}. For token merging, PatchMerger \cite{renggli2022learning} proposes a module that reduces the number of tokens by merging them between two consecutive intermediate layers \cite{renggli2022learning}. Recently, a hybrid solution, Token Fusion ({ToFu}) \cite{Kim_2024_WACV}, proposed to put together the benefits of both token pruning and token merging. Token selection can also be combined with the other methods discussed in this paper. As an example, Adaptive Computation Module (ACM) \cite{wojcik2023adaptive} is a technique that combines token dropping with a dynamic width principle, in which each token is allocated a variable width for each layer in the network.

\begin{table*}[]
\caption{{Overview of methods to improve efficiency of the models, including baseline algorithms not discussed in the text (e.g., pruning). \textbf{Parameters}: whether the method adds additional parameters. \textbf{Pre-training}: whether the method requires a pre-training phase. \textbf{Latency/FLOPs}: whether the method optimizes one of these two metrics}}\vspace{1em}
\label{tab:eff}
\footnotesize
\begin{tabular}{cccccc}
\hline
Method         &   Approach      & Parameters & Pre-training & Latency &  FLOPs \\ \hline
Dynamic Convolutions \cite{Verelst_2020}  (2020)               &  Sparsity        &       Yes       &      -          &     Yes    &  Yes   \\
Scaling Transformers \cite{jaszczur2021sparse}   (2021)         &      Sparsity    &       -       &        -        &    Yes     &  -   \\
SViTE \cite{NEURIPS2021_a61f27ab}  (2021) &     Pruning         &         -     &        -        &   Yes      &   Yes  \\
Sparse Momentum \cite{dettmers2019sparse}   (2019)         &       Pruning   &        -      &       -         &   Yes   &   Yes     \\
SparseViT \cite{Chen_2023_CVPR}    (2023)          &    Pruning      &       Yes       &      -          &    Yes      &  Yes  \\
Sparse Transformers \cite{child2019generating}    (2019)       &   Factorization       &     Yes         &       -         &    -    &    Yes  \\
Differentiable branching \cite{scardapane2020differentiable} (2020)  &     Early Exits     &     Yes         &       -         &  Yes       &  Yes   \\
Adaptive Early Exits \cite{bolukbasi2017adaptive}     (2017)    &    Early Exits      &    Yes          &      -          &      Yes  &  -    \\
L2W-DEN \cite{han2022learning}   (2022)         &      Early Exits    &        Yes      &        -        &    Yes    &    -  \\
PABEE \cite{zhou2020bert}      (2020)     &    Early Exits      &     Yes         &       -        &   Yes     &    -  \\
Branchynet \cite{teerapittayanon2016branchynet} (2016) &         Early Exits &      Yes        &         -       &   Yes     &    -  \\
AEP \cite{sarti2023}      (2023)       &     Early Exits     &       Yes       &     -           &     Yes    &   Yes  \\
BoF \cite{passalis2020}     (2020)             &   Early Exits       &     Yes         &          -      &     -     &    Yes \\
DeepSpeed-MoE\cite{rajbhandari2022deepspeedmoe} (2022)  &   MoE       &        Yes      &       Yes         &   Yes    &   Yes    \\
Sparsely-Gated MoE \cite{shazeer2017outrageously}   (2017)    &   MoE       &     Yes         &      -          &    -     &  Yes   \\
V-MoE \cite{riquelme2021scaling}   (2021)        &    MoE      &         Yes     &        -        &    -    &   Yes   \\
MoA \cite{zhang2022mixture}  (2022)            &   MoE       &       Yes       &      -          &      -    &  Yes  \\
SUT \cite{tan2023sparse}  (2023)         &     MoE     &          Yes    &       -         &     -     &   Yes \\
MoV/MoLoRA \cite{zadouri2023pushing}   (2023)         &   MoE       &         Yes     &      Yes          &    -    &   Yes   \\
EMoE \cite{qiu2023emergent}    (2023)           &    MoE      &       -       &      Yes          &    -    &   Yes   \\
GMoE \cite{li2022sparse}     (2022)          &   MoE       &          Yes    &       Yes         &   -     &    Yes  \\
MoEfication \cite{zhang2021moefication}          &    MoE      &       Yes       &         Yes       &    -   &    Yes   \\
SADMoE \cite{piorczynski2023exploiting} (2023)    &     MoE     &       Yes       &            Yes    &    -    &  -    \\
A-ViT \cite{yin2022vit}    (2022)          &     Token Sampling     &       -       &       -         &    Yes      &  Yes  \\
ToMe \cite{bolya2022token, bolya2023token}    (2022)            &    Token Sampling      &      Yes        &        -        &     Yes   &   Yes   \\
AdaViT \cite{meng2022adavit}       (2022)         &    Token Sampling      &       Yes       &      -          &    -     &     Yes \\
DynamicViT \cite{rao2021dynamicvit}   (2021)          &   Token Sampling       &       Yes       &          -      &    Yes   &   Yes    \\
MSViT \cite{Havtorn_2023_ICCV}    (2023)       &     Token Sampling     &       Yes       &       -         &    Yes      &  Yes  \\
GTP-ViT \cite{Xu_2024_WACV}      (2024)         &    Token Sampling      &    Yes          &      -          &    Yes     &   Yes  \\
Colt5 \cite{ainslie2023colt5}      (2023)        &    MoE      &       Yes       &       Yes         &    Yes     &   Yes  \\
PatchMerger \cite{renggli2022learning}   (2022)       &     Token Sampling     &      Yes        &         Yes       &   Yes   &     Yes   \\
ToFu \cite{Kim_2024_WACV}    (2024)           &    Token Sampling      &     -         &     Yes           &   Yes   &     Yes   \\
ACM \cite{wojcik2023adaptive}     (2023)       &   MoE/Token Sampling       &          Yes    &       Yes         &  -   &   Yes     \\ \hline
\end{tabular}
\end{table*}

\subsection{Specialization}
\label{subsec:specialization}

Modularity can also lead to specialization benefits in specific tasks \cite{zhang2023emergent, mittal2022modular}, such as language modeling \cite{lin-etal-2021-learning, mialon2023augmented}, cross-lingual transfer \cite{pfeiffer-etal-2022-lifting, ansell-etal-2022-composable, pfeiffer-etal-2022-xgqa, vu-etal-2022-overcoming}, speech processing \cite{mialon2023augmented, pfeiffer-etal-2021-adapterfusion}, computer vision and multi-modality \cite{rajendran2017attend, mustafa2022multimodal, 7780381}, and task generalization \cite{muqeeth2024learning}. 

Specialization can happen in two ways: \textit{implicitly} if no external information is provided, or \textit{explicitly}  if additional information (e.g., the speaker identity) is known. The key insight is that knowledge of, e.g., the task, the domain, or the speaker provides latent information that can be used to condition the routing blocks ($\Gamma$ in Sec. \ref{sec:formulation}), thus specializing specific parts of the network or specific components to different scenarios. This can be achieved in different ways: manual routing in which different components are pre-selected for the different latent vectors \cite{pfeiffer2023modular}, direct conditioning of the routing functions \cite{zhu2022uni}, entropy regularizers to force the distributions w.r.t. to a specific latent vector to have low entropy \cite{mustafa2022multimodal}, or weight sharing across tasks \cite{tang2023you}. Soft merging \cite{muqeeth2023soft} also seems to offer specialization benefits.

Studies on quantifying the quality of the resulting specializations were carried out recently in the literature \cite{xue2024openmoe}.
MoEs showed promising performance due to their nature, and, specifically, hard routing facilitates module specialization \cite{pfeiffer2023modular}. On the other hand, learned routing could lead to sub-optimal results \cite{mohammed2022models, mittal2022modular} w.r.t to fixed routing, except for specific cases \cite{ponti2022combining}. As another example, \cite{xue2024openmoe} showed that many routing decisions tend to ignore context, and they can be fixed during the early stages of training. Generally speaking, understanding the interplay between routing and specialization and the extent to which this specialization correlates with human-understandable semantics and biological plausibility remain open challenges that will require novel benchmarks and metrics \cite{qiu2023emergent,zhang2023emergent}. Recent moefications {works} show that some form of emergent modularity may also be found in pre-trained networks with no specific modularity bias \cite{qiu2023emergent,zhang2021moefication}.

A benefit of having specialized sub-structures and components is that the networks can perform better in multi-task learning \cite{muqeeth2024learning} and multi-domain scenarios \cite{mustafa2022multimodal}, and they can potentially enable zero-shot transfer and generalization of the resulting sub-structures. We list a few interesting examples from the recent literature. PHATGOOSE \cite{muqeeth2024learning} is a MoE specialized in zero-shot generalization, thanks to a new post-hoc routing strategy. EMoE focuses on the implicit modular structures (Emergent Modularity) of large pre-trained transformers \cite{qiu2023emergent}. DSelect-k \cite{hazimeh2021dselect} presents a continuously differentiable and sparse gate for multitask learning. Uni-Perceiver-MoE \cite{zhu2022uni} is a generalist conditional MoE that shows SOTA performance when compared to specialized MoEs. LIMoE \cite{mustafa2022multimodal} is focused on language-image pre-training. DeepSeekMoE \cite{dai2024deepseekmoe} manages to ensure expert specialization for language models, still reducing computational costs. Other approaches include modular submodels to scale language models \cite{Biadsy_2022} and a class-aware channel pruning for \textit{queriable} NNs \cite{JIN2022186}.

\subsection{Explainability}
\label{subsec:explainability}

Finally, modularity and sparsity can provide significant gains in explainability, which is a significant issue when deploying systems \cite{hassija2024interpreting}. In particular, analyzing and plotting the routing decisions made by the subsampling blocks $\Gamma$ almost always provide valuable insights into the predictions. These include, but are not limited to, visualizing representative tokens sent to each expert \cite{mustafa2022multimodal}, visualizing the early exits distribution for a given sequence in an autoregressive model \cite{schuster2022confident}, visualizing the tokens that were never discarded in a network having token selection \cite{NEURIPS2021_d072677d, meng2022adavit, rao2021dynamicvit}, or the number of experts that were activated token-wise \cite{wojcik2023adaptive}. These techniques can provide benefit also for specific tasks, such as object detection \cite{Chen_2023_CVPR}. 

Also in this case, we lack principled benchmarks and frameworks to analyze the resulting plots, which is an open problem in explainability in general \cite{hassija2024interpreting}. In addition, for networks having thousands of blocks or experts, manually analyzing each of them can be a time-consuming process. LLMs can potentially help in automating this process \cite{bills2023language}. From a mechanistic interpretability point of view, fully understanding and being able to track the evolution of modules in these networks could be a large step forward in better understanding overfitting, generalization, and fine-tuning \cite{huang2024unified}. Finally, we note that being able to plot and visualize internal decisions of the network provides a direct way for providing feedback on, e.g., whether specific modules should be kept or ignored, and to perform interventional analyses \cite{bontempelli2022concept}.

\section{Conclusions and future trends}
\label{sec:conclusions}

In this tutorial paper we have provided an introduction to the emerging field of designing neural networks which are \textit{sparsely activated} in a \textit{modular} fashion, via the use of conditional computation techniques. To this end, we have provided both a general mathematical formalism to categorize these approaches, and then an overview of several concrete implementations including mixture-of-expert models and early exit neural networks. Although these models have been investigated mostly for improving training and/or inference time, we have discussed a number of additional emerging benefits from this approach, including specialization, generalization, and explainability. Many of these benefits are only starting to be investigated, opening up interesting avenues of research.

Some common challenges have also emerged from our discussion: (a) being able to control and adapt the inference and/or training time is still an open challenge, with most techniques providing a fixed accuracy-time trade-off (e.g., MoEs with top-$k$ routing); (b) sparse routing introduces balancing and collapse challenges that must be taken care of; (c) most routing decisions are taken only locally (e.g., layer-wise), while taking them globally requires sampling from a combinatorially large search space requiring new sampling techniques \cite{niculae2023discrete}; (d) outside of accuracy, benchmarks and metrics for evaluating specialization and generalization of these models are still being developed \cite{mittal2022modular}.

There are also several research directions that we were not able to touch due to space constraints: these include scaling laws for modular models \cite{clark2022unified}, biological plausibility \cite{scardapane2020should,zhang2023emergent}, and exploiting these models in specific applicative fields such as split computing \cite{matsubara2022split} and semantic communication \cite{zhang2022unified}.

\section*{Acknowledgments}
S. Scardapane is partly funded by Sapienza grants RM1221816BD028D6 (DESMOS) and RG123188B3EF6A80 (CENTS).
P. Minervini is partially funded by ELIAI (The Edinburgh Laboratory for Integrated Artificial Intelligence), EPSRC (grant no. EP/W002876/1), an industry grant from Cisco, and a donation from Accenture LLP.

\bibliographystyle{abbrv}
\bibliography{biblio.bib}

\end{document}